\documentclass[conference]{IEEEtran}
\ifCLASSINFOpdf
\else
\fi
\usepackage[square, comma, sort&compress, numbers]{natbib}
\usepackage{graphicx} 
\usepackage{amsfonts}
\usepackage{amsmath}
\usepackage{multirow}
\usepackage{booktabs} 
\usepackage{array}
\usepackage{algpseudocode}
\usepackage{amsmath}
\usepackage{bm}
\usepackage{amsmath}
\usepackage{url} 
\usepackage{booktabs} 
\usepackage{stfloats} 
\usepackage{xcolor}
\usepackage[linesnumbered,ruled,vlined]{algorithm2e}
\usepackage[colorlinks,linkcolor=red,citecolor=black]{hyperref} 

\SetCommentSty{mycommfont}
\SetKwInput{KwInput}{Input}                
\SetKwInput{KwOutput}{Output}

\begin{document}
%
\title{Fooling the Eyes of Autonomous Vehicles: Robust Physical Adversarial Examples Against Traffic Sign Recognition Systems}

\author{\IEEEauthorblockN{Wei Jia}
\IEEEauthorblockA{School of Cyber Science and Engineering\\Huazhong Univ. of Sci. \& Tech.\\
jiaw@hust.edu.cn}
\and
\IEEEauthorblockN{Zhaojun Lu*}
\IEEEauthorblockA{School of Cyber Science and Engineering\\Huazhong Univ. of Sci. \& Tech.\\
lzj\_cse@hust.edu.cn}
\and
\IEEEauthorblockN{Haichun Zhang}
\IEEEauthorblockA{Huazhong Univ. of Sci. \& Tech.\\
homer@thesimpsons.com}
\and
\IEEEauthorblockN{Zhenglin Liu}
\IEEEauthorblockA{Huazhong Uni. of Sci. \& Tech.\\
liuzhenglin@hust.edu.cn}
\and
\IEEEauthorblockN{Jie Wang}
\IEEEauthorblockA{Shenzhen Kaiyuan Internet Security Co., Ltdy\\
wangjie@seczone.cn}
\and
\IEEEauthorblockN{Gang Qu}
\IEEEauthorblockA{University of Maryland\\
gangqu@umd.edu}
}


%


\IEEEoverridecommandlockouts
\makeatletter\def\@IEEEpubidpullup{6.5\baselineskip}\makeatother
\IEEEpubid{\parbox{\columnwidth}{
}
\hspace{\columnsep}\makebox[\columnwidth]{}}

\maketitle

\begin{abstract}
Adversarial Examples (AEs) can deceive Deep Neural Networks (DNNs) and have received a lot of attention recently. However, majority of the research on AEs is in the digital domain and the adversarial patches are static. Such research is very different from many real-world DNN applications such as Traffic Sign Recognition (TSR) systems in autonomous vehicles. In TSR systems, object detectors use DNNs to process streaming video in real time. From the view of object detectors, the traffic sign’s position and quality of the video are continuously changing, rendering the digital AEs ineffective in the physical world.

In this paper, we propose a systematic pipeline to generate robust physical AEs against real-world object detectors. Robustness is achieved in three ways. First, we simulate the in-vehicle cameras by extending the distribution of image transformations with the blur transformation and the resolution transformation. Second, we design the single and multiple bounding boxes filters to improve the efficiency of the perturbation training. Third, we consider four representative attack vectors, namely Hiding Attack (HA), Appearance Attack (AA), Non-Target Attack (NTA) and Target Attack (TA). For each of them, a loss function is defined to minimize the impact of the fabrication process on the physical AEs.

We perform a comprehensive set of experiments under a variety of environmental conditions by varying the distance from $0m$ to $30m$, changing the angle from $-60^{\circ}$ to $60^{\circ}$, and considering illuminations in sunny and cloudy weather as well as at night. The experimental results show that the physical AEs generated from our pipeline are effective and robust when attacking the YOLO v5 based TSR system. The attacks have good transferability and can deceive other state-of-the-art object detectors. We launched HA and NTA on a brand-new 2021 model vehicle. Both attacks are successful in fooling the TSR system, which could be a lifethreatening case for autonomous vehicles. Finally, we discuss three defense mechanisms based on image preprocessing, AEs detection, and model enhancing.

\end{abstract}


%

\section{Introduction}

Artificial Intelligence (AI) and Deep Neural Networks (DNNs) have boosted the performance of a large variety of applications, in particular computer vision tasks such as face recognition \cite{R1}, image classification \cite{R2}, and object detection \cite{R3}. Unfortunately, the ubiquity and diversity of AI applications have also created incentives and opportunities for attackers to attack DNNs for malicious purposes \cite{R4}. In 2014, Szegedy \emph{et al.} \cite{R5} first introduced the concept of Adversarial Examples (AEs), which are well-crafted examples with imperceptive perturbations that deceive the image classifiers into misclassification. A lot of follow-up studies \cite{R6,R7,R8,R9,R10} demonstrated the vulnerabilities of DNNs under different types of AEs. One common feature of these attacks is that they are all in the digital domain instead of physical domain. Compared to the theoretical studies which have yielded prolific results on AEs and led to better understanding the principles of DNNs, the practicality of these AEs and their impact on real-world AI applications remain under-investigated. 

Recently there have been reported studies on the feasibility of AEs in the physical domain, where the digital adversarial images are printed to the physical domain first, and then pictures are taken from these physical images to create AEs \cite{R11,R12,R13,R14,R15}. However, the physical AEs generated by this digital-physical-digital conversion becomes significantly less effective because of complicated physical conditions during the process such as the distance, angle, and illumination when the images are re-taken. Consequently, there are approaches to improve the robustness. Initial efforts have been devoted to improving the robustness of AEs. Athalye \emph{et al.} \cite{R12} introduced the Expectation Over Transformation (EOT) method to simulate the effect of rotation, scaling, and perspective changes. Evtimov \emph{et al.} \cite{R13} and Sitawarin \emph{et al.} \cite{R14} enhanced this method by synthesizing the adversarial traffic signs to attack the image classifiers. Object detectors are adopted for the Traffic Sign Recognition (TSR) task to assist the safety- and security-critical autonomous driving systems \cite{R15}.

Unlike image classifiers, fooling object detectors with physical AEs is much more challenging for the following reasons. Firstly, image classifiers only process static images, but the object detectors like those in autonomous vehicles are commonly working in environments where physical features of the object such as its relative position to the object detector keep changing. Secondly, the imperfections in the fabrication process have an uncontrollable impact on the effectiveness of AEs, creating a large gap between the digital domain and the physical domain for adversarial attacks. The digital-physical-digital conversion would weaken the toxicity of the fabricated AEs. Thirdly, the AEs generation algorithms require some information of the targeted DNNs, which may not be realistic because attackers may not have control over the built-in systems of autonomous vehicles. One practical approach is to generate AEs under the white-box settings and use them to attack the black-box object detectors. To summarize, \textit{the key to fooling object detectors, the eyes of autonomous vehicles, is to improve the robustness of physical AEs}. 

In this paper, we propose a systematic pipeline to generate robust physical AEs and demonstrate its effectiveness against the object detectors used in TSR systems. To reflect the real-world scenarios, we generate physical AEs for traffic signs with a large range of physical parameters including distance, angle, and illumination. First, we extend the distribution of image transformations with the blur transformation and the resolution transformation to simulate the in-vehicle cameras. Then, Single Bounding Box (S-BBOX) and Multiple Bounding Box (M-BBOX) filters are designed to obtain the relative BBOXes before generating the physical AEs. We define four loss functions to train adversarial perturbations corresponding to the following four attack vectors: Hiding Attack (HA), Appearance Attack (AA), Non-Target Attack (NTA), and Target Attack (TA). HA hides AEs in the background so that the object detectors cannot detect them. AA makes the object detectors recognize a bizarre AE as a common category. Both NTA and TA deceive the object detector into misrecognition with imperceptible AEs. TA is more destructive since it makes the object detectors recognize an AE of one category as an object from another designated category. To improve the transferability of AEs, we use different background images in the AEs generation algorithm to avoid overfitting.

Finally, we validate the robustness of our physical AEs by driving a brand-new 2021 model vehicle toward the physical AEs to see whether the vehicle's TSR system will be fooled. Our main contributions can be summarized as follows: 

\subsubsection{\textbf{A systematic approach to generate robust physical AEs}}  In our physical AE generation pipeline, we propose several approaches to improve the robustness of the physical AEs. We extend the distribution of image transformations to simulate the complicated environmental driving conditions. S-BBOX and M-BBOX filters are designed to obtain the BBOXes associated with the target object to train perturbation efficiently. Four loss functions are defined to generate AEs for attack vectors.

\subsubsection{\textbf{A comprehensive set of experiments}} To systematically evaluate the effectiveness and robustness of the generated physical AEs, we conduct extensive outdoor experiments. More than 1,000 video clips containing more than 100,000 image frames are taken by a high-resolution camera. Real driving scenarios are simulated with varying distances from $1m$ to $30m$, angles from $-60^{\circ}$ to $60^{\circ}$, and illuminations for sunny, cloudy, and night.

\subsubsection{\textbf{Successful attacks against YOLO v5 based object detector and TSR system in a 2021 model vehicle}} To the best of our knowledge, this is the first set of adversarial attacks against YOLO v5 based object detectors in the physical domain. We successfully launch four attack vectors, especially NTA and TA, that are life-threatening in the real world. Our physical AEs also exhibit satisfactory transferability when attacking a production-grade TSR system of a brand-new 2021 model vehicle.

\section{Background}

In this section, we first introduce the advances in the field of image classification and object detection and explain why autonomous vehicles adopt object detectors for the TSR task. Then, we summarize the difficulties and limitations of the existing physical adversarial attacks against the object detectors.  

\subsection{Object Detection}

An image classifier $\bm{f}(\cdot): \mathbb{R}^{h\times w\times\varepsilon} \rightarrow \mathbb{R}^N$ recognizes an input image $x\in \mathbb{R}^{h\times w\times\varepsilon}$ to a category label $y_x \in (1, ..., N)$ where $h$ ,$w$ and $\varepsilon$ is the height, width and channel of $x$. The image classifier is commonly trained with supervised learning strategy, whose goal is to minimize a loss function between the output of the network $\bm{f}(x)$ and the expected label $y_x$ \cite{R17}. However, image classification cannot satisfy the demands of processing dynamic video streaming containing multiple objects for autonomous vehicles \cite{R18}. Take the TSR task for example, it is vital for the vehicle sensor to recognize a traffic sign quickly and accurately from the complicated road background, then give the correct instruction to the vehicle controller for rapid response \cite{R19}. Object detection based on DNNs is adopted by autonomous vehicles to process consecutive frames of images containing multi-category objects (\emph{e.g.}, traffic signs, vehicles, pedestrians, and cyclists) \cite{R20}. Modern DNNs based object detectors can be classified into two categories: one-stage architecture represented by YOLO series with higher detecting speed \cite{R21}, and two-stage architecture represented by Faster R-CNN with higher detecting accuracy \cite{R22}. Since one-stage architecture processes BBOX regression and object classification concurrently without a region proposal stage, it is much faster than two-stage architecture so that meets the real-time requirement \cite{R23}. In recent years, many algorithms have been developed to balance the speed and the accuracy of the object detectors. Bochkovskiy \emph{et al.} \cite{R24} proposed YOLO v4 that improved the accuracy compared with YOLO v3 \cite{R25} while maintaining the real-time object detection capability. Soon after the release of YOLO v4, a company named Ultralytics released YOLO v5's source code at Github \cite{R26}. YOLO v5 has higher mean Average Precision (mAP) and lower processing time than YOLO v4. It should be emphasized that YOLO v5 is trained with clear images, but it can be used to detect blurred images \cite{R16}. The prominent features allow YOLO v5 to detect objects under various conditions, which is well-performed in the TSR task.

\begin{figure}[t]
  \centering
  \includegraphics[width=\linewidth]{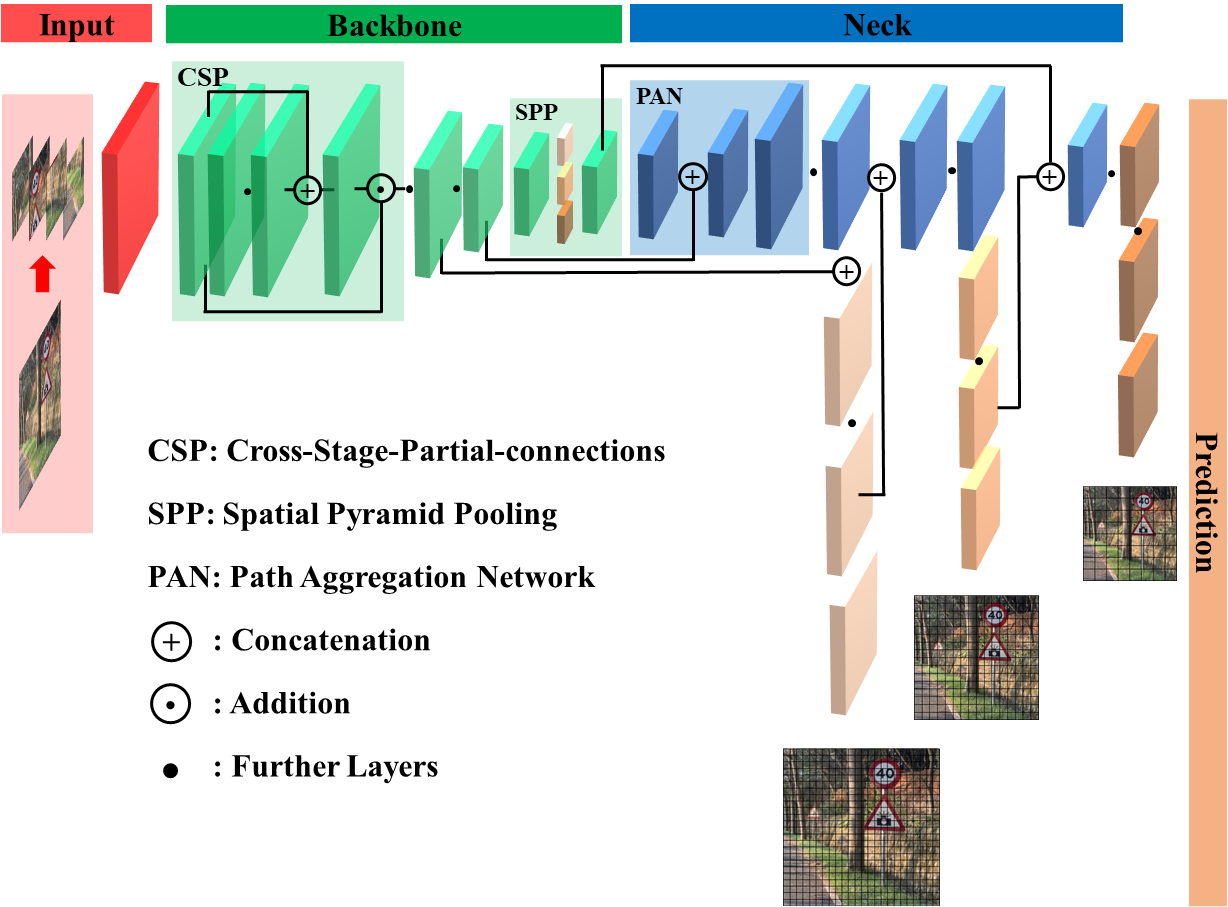}
  \caption{Architecture of YOLO v5.}
\end{figure}

As briefly introduced in Fig.1, YOLO v5 inherits some advanced techniques from YOLO v4, including the Darknet53 with Cross Stage Partial \cite{R27} (CSPDarknet53) as backbone, the Path Aggregation Network \cite{R28} (PANet) as neck, and the Spatial Pyramid Pooling \cite{R29} (SPP) block to increase the receptive field. Besides, YOLO v5 makes some fine-tuning on the basis of YOLO v4 such as slicing input images into four small ones and  concatenating them together for convolution operation. As a classical anchor-base object detector, YOLO v5 first pre-processes the anchors, then corrects the Bounding Box (BBOX) according to the off-set values, finally obtains the probability of the detected object in this BBOX \cite{R24}. Each anchor outputs three parts, the object probability refers to the possibility that an object exists in this anchor, off-set value of BBOX refers to the off-set between the anchor BBOX and the ground truth BBOX, and the probability vector of all categories refers to the possibility that the object in this anchor belongs to each category. The number of anchors depends on the size of the feature map, in which each point has three anchors with different shapes predetermined by hyper-parameters. We have trained and tested the state-of-the-art object detectors on the TT100k dataset \cite{R30}. YOLO v5 achieves the best performance in terms of real time and accuracy. Therefore, we chose the YOLO v5 based TSR system as the attack target to evaluate our proposed physical adversarial attack.

\subsection{Physical Adversarial Examples}

Most of the prior research either focused on the digital adversarial attacks, or launched physical adversarial attacks against the image classifiers \cite{R31}. However, successfully attacking a few static images cannot threaten the object detectors dealing with video streaming. There is no practical and satisfactory attempt to physically attack the object detectors due to three major challenges, \emph{i.e.}, cross-domain conversion, image transformation, and limited capability. 

\subsubsection{Cross-Domain Conversion}

As illustrated in Fig.2, AEs are generated in the digital domain, then printed into the physical domain, and finally re-taken by the camera back to the digital domain. The experimental results show that the digital-physical-digital conversion significantly degrades the adversarial toxicity of AEs \cite{R32}. To address this challenge, Jan \emph{et al.} \cite{R33} presented an image-to-image translation network to simulate the digital-physical conversation. A conditional Generative Adversarial Network (cGAN) \cite{R34} is used to learn the digital-physical conversion for generating a synthetic physical image, then it is served to produce adversarial noises in AEs generation. The cGANs model needs to be trained with a set of paired static images. Thus, the robustness of the physical AEs is improved when attacking several image classifiers.

\subsubsection{Image Transformation}

Different distances, angles, and illuminations will result in image transformations that impact the robustness of the physical AEs. AEs generated through the L-BFGS attack \cite{R5}, fast gradient sign method \cite{R7}, and the C\&W attack \cite{R6} often lose their adversarial nature once subjected to minor transformations \cite{R35,R36}. To address this challenge, Athalye \emph{et al.} \cite{R12} introduced the Expectation Over Transformation (EOT). EOT uses a chosen distribution of transformation functions and constrains the expected effective distance between AEs and the input images. EOT is able to generate 3D printed AEs which remain adversarial under a range of conditions. Within the framework of EOT, the physical AEs successfully deceive a standard image classifier.

\subsubsection{Limited Capability}

Further difficulty comes from the fact that the DNNs model is usually only a component in the whole computer vision system. For most applications, the attackers can neither get access to data inside the system nor obtain the parameters of the DNNs model. Instead, they can only manipulate objects in the physical environment. Therefore, the limitation of the attacker's capability limits the threat of the physical AEs. NaturalAE \cite{R37} proposed in 2021 used the natural AEs generated under white-box settings to attack the black-box models. However, the attack success rate is low, and the attack range is only $5m$. For a moving vehicle, the attack is only effective within less than half a second.

\begin{figure}[t]
  \centering
  \includegraphics[width=\linewidth]{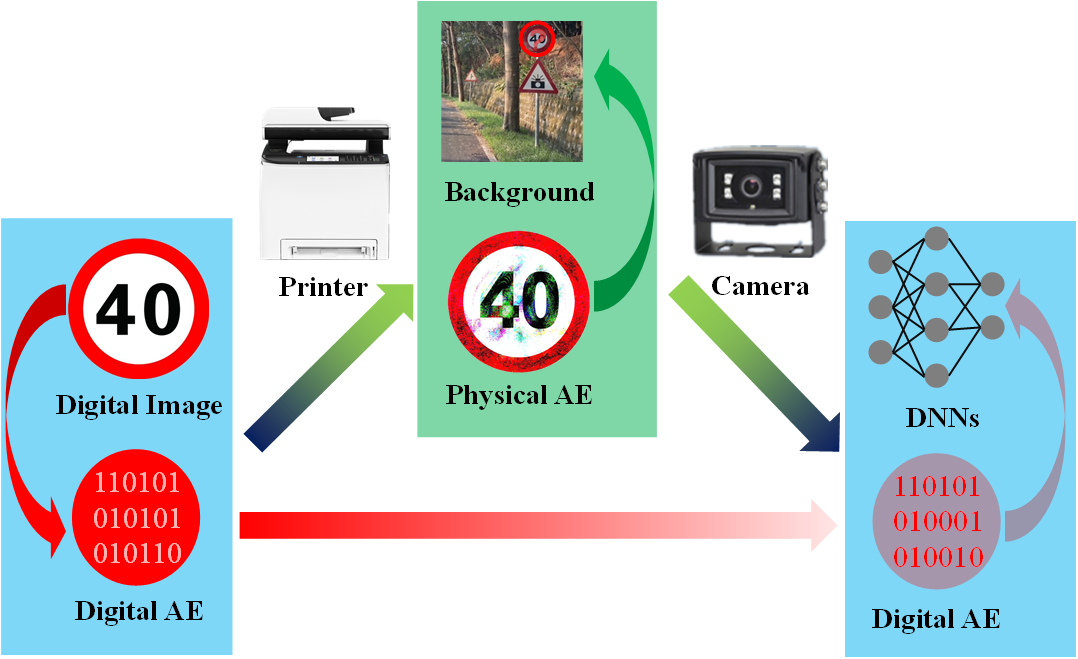}
  \caption{The Digital-Physical-Digital cross-domain conversion of AEs.}
\end{figure}

\section{The Proposed Physical AE Pipeline}

In this section, we will elaborate on our proposed physical AEs pipeline. We first give the threat model with the goal and capabilities of the attacker. Then we present the 3-step AE attack pipeline which addresses the challenges in the previous section. Finally, we describe each of the three steps and show in detail the AEs generation algorithms for four attacks: Hiding Attack (HA), Appearance Attack (AA), Non-Target Attack (NTA), and Target Attack (TA).


\subsection{Threat Model}

\begin{figure}[t]
  \centering
  \includegraphics[width=\linewidth]{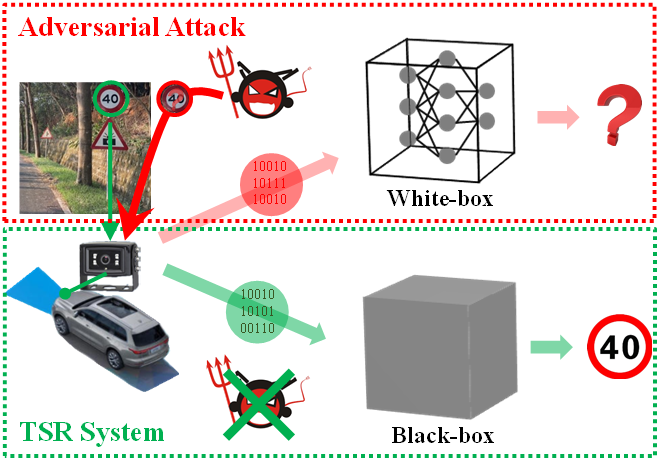}
  \caption{Threat model of physical adversarial attack: the TSR system is a black box and the attacker does not have access to it.}
\end{figure}

\begin{figure*}[t]
  \centering
  \includegraphics[width=\linewidth]{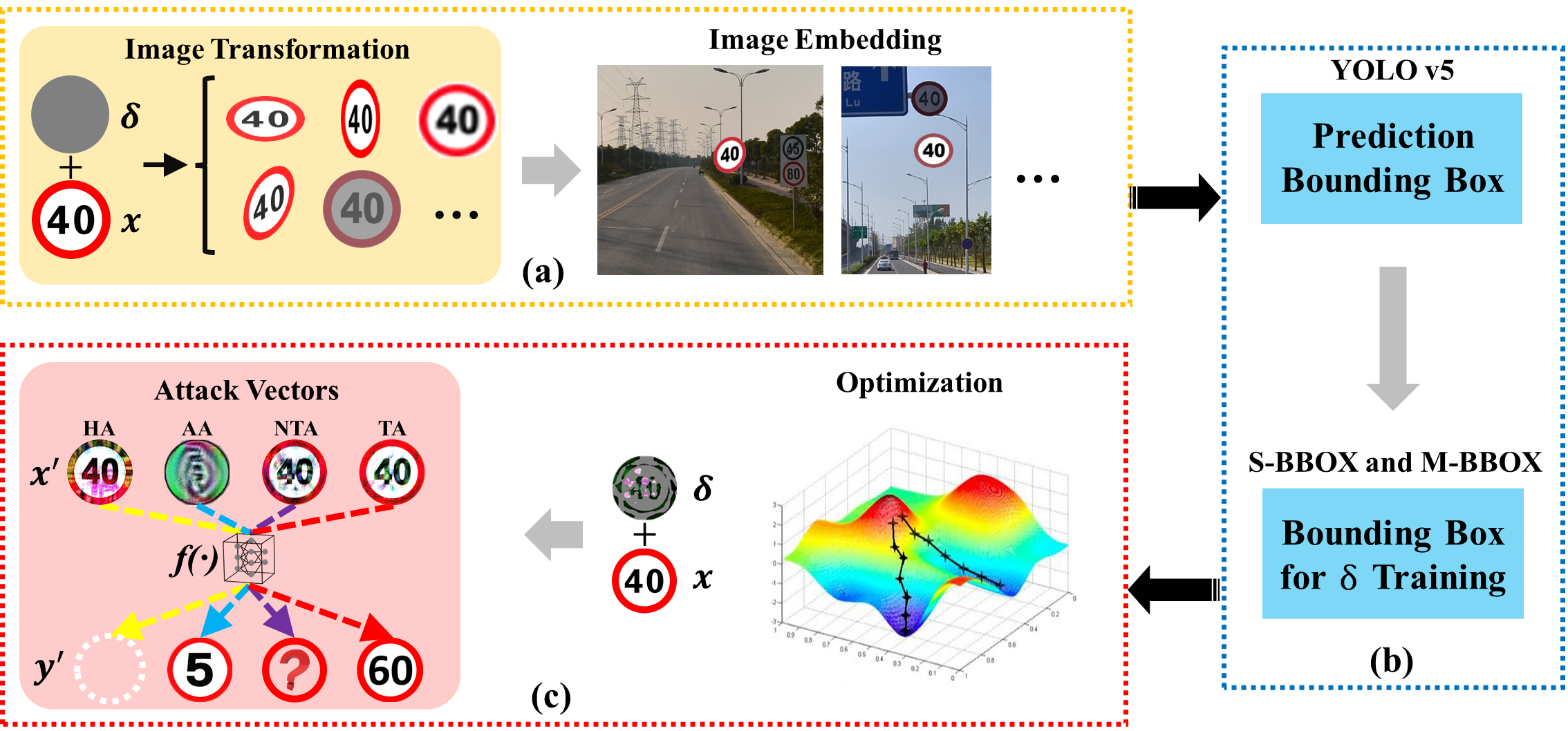}
  \caption{Physical adversarial attack pipeline. (a) Image transformation and embedding. (b) BBOX filter. (c) Perturbation training for four attack vectors.}
\end{figure*}

The attacker's goal is to mislead the victim vehicle's TSR system to incorrect classification of a given traffic sign. To this end, the attack can be further modeled as Hiding Attack (HA) where the attacker wants to hide the traffic sign, Appearance Attack (AA) where the attacker wants to create an object which can be recognized as a specific sign, Non-Target Attack (NTA) where the attacker simply wants the classification result to be incorrect, and Target Attack (TA) where the attacker wants the misclassified result to be a specific sign that is not the original traffic sign. Formally speaking, let $\bm{f}(\cdot)$ be the image classifier used in the TSR system, for a given input image $x$ whose category label is $y$, the attacker wants the TSR system to output a label that is different from $y$. 

We adopt (i) the black-box TSR model where the attacker does not know the implementation details of the TSR system such as its structure and parameters. Furthermore, we assume that (ii) the attacker does not have physical access to the in-vehicle networks to analyze or modify the data in the TSR system. However, the attacker (iii) has access to the traffic sign which he can physically modify. Under these assumptions, the attacker's goal can be formulated as, generate an AE $x' = x + \delta$ for the input image $x$, where $\delta$ is a perturbation to $x$, such that:

\begin{equation}
\bm{f}(x') \neq y
\end{equation}

We believe that this is the most natural and strongest model for the attack scenario where an attacker wants to fool the TSR system of a victim vehicle. Any prior knowledge of the victim vehicle's information such as the model of the TSR, the object detector it uses, the camera(s) equipped in the vehicle, and the driving status (\emph{e.g.} speed, weather, road condition) might provide additional help to the attacker and make the attack easier. 

It is important to clarify that the previous AEs in the digital domain cannot be applied to this threat model. First, adversarial attack assumes a white-box model of the target DNN, the TSR system is a black-box and the parameters and structure of the target DNN models used in the TSR (assumption i) are not available to the attacker. So the attacker will not be able to generate digital AEs directly (see the top of Fig.3). Second, even if the attacker manages to create some digital AEs, because he does not have access to the in-vehicle networks (assumption ii), the attacker will not be able to inject such digital AEs directly into the TSR system (see bottom of Fig.3). However, the attacker can modify the traffic sign and hope the victim vehicle's TSR system will misclassify it.



\subsection{Attack Pipeline}

As illustrated in Fig.4, there are three major steps in the proposed adversarial attack pipeline.

\textbf{Step 1.} As shown in Fig.4(a), in order to improve the robustness of the physical AEs, we extend the distribution of image transformations to simulate the changes of distance, angle, and illumination in the real world. The transformed images will be embedded into the background as the foreground to simulate the perspective of the object detectors.

\textbf{Step 2.} As shown in Fig.4(b), The BBOXes are associated with $x$, and can be extracted from the prediction results of YOLO v5. Single BBOX (S-BBOX) filter and Multiple BBOX (M-BBOX) filter are designed to obtain BBOXes for efficient perturbation training.

\textbf{Step 3.} As shown in Fig.4(c), each BBOX contains three types of information, \emph{i.e.}, the probability of containing $x$ or $x'$, the probability vector for each category, and the BBOXes position offset. With the first two pieces of information, four loss functions are presented to generate AEs for aforementioned four attack vectors respectively.

\subsection{Image Preprocessing}

The conventional AEs generation algorithms adopted the image transformations in EOT \cite{R12}, including rotation, perspective, brightness, contrast, saturation, hue, and Gaussian noise. Furthermore, we extend distribution with the blur transformation to simulate conditions such as camera shake, and the resolution transformation that improves the robustness of AEs for varying distance to the camera. In step 1, we address the image transformation by setting:

\begin{equation}
x_{t} = \bm{t}(\bm{M}(x'))
\end{equation}
\noindent
where $\bm{M}(\cdot)$ is the mask function to constrain the area where the perturbation is added. $\bm{t}(\cdot)$ is a transformation vector randomly selected from distribution of image transformations $T$. 

\begin{algorithm}[t]
\DontPrintSemicolon
\caption{Image Embedding}
    \KwInput{$r_1$,$r_2$,$r_3$: three ratios of foreground; \qquad \qquad $i$: scale of background image.}
    \KwOutput{$bbox_{real}$: bounding box information of foreground.}
    \If{$\bm{random()}>0.8$} 
    {
        $m = r_2-r_1$\;
        $a = r_1$\;
    }
    \Else 
    {
        $m = r_3-r_2$\;
        $a = r_2$\;
    }
    $w_{real} = h_{real} = \bm{random()} \times m+a$\; 

    $s_f = \frac{3}{i}$\; 
    $p_x = \bm{random()} \times (1-w_{real}-2 \times s_f )+0.5 \times w_{real}+s_f$        \tcp*{avoid clinging the edge}
    $p_y = \bm{random()} \times (1-h_{real}-2 \times s_f )+0.5 \times h_{real}+s_f$\;
    $bbox_{real} = [p_x , p_y , h_{real} , w_{real}] \times i$\; 
\end{algorithm}

We have attempted to fix the center point of the foreground at any position in the entire background. However, since part of the foreground is out of the background, it will make the perturbation training difficult to converge. Considering the fact that the camera will capture the complete traffic signs in most cases, we keep a certain distance between the center point and the edge of the background to guarantee that the entire foreground can be processed in the AEs generation algorithm. The image embedding algorithm is shown in Algorithm 1. During the optimization procedure, each transformed image should be embedded in different traffic backgrounds to imitate the perspective of the object detectors. $\bm{random}()$ can output a random value in $[0,1]^\mathbb{R} $ every time it is called. $(p_x,p_y)$, $h_{real}$ and $w_{real}$ are the center point, height and width of the foreground respectively. $s_f$ is a position factor to prevent the foreground from clinging to the edge of background or even out of background. To simulate the real scale change of foreground in the background, we choose two different scale ranges corresponding to the big object and small object to generate the height and width of each BBOX. $r_1$, $r_2$, and $r_3$ indicate the ratio of the transformed images embedded in the background. The higher the ratio, the larger the object. We found that when the object is small, the direction of gradient update is hard to search and the model may not converge. However, if the object is far from the camera, lowering the ratio will enhance the effectiveness of AEs. Our empirical study shows that the best results come from when we choose the ratio in [0.01, 0.1] with probability 20\% and in [0.1, 0.5] with 80\%. Therefore, $(r_1,r_2,r_3)$ is set to $(0.01,0.1,0.5)$ in our experiment.

\subsection{Bounding Box Filter}

As mentioned before, YOLO v5 detects the object according to the feature maps of three different scales. For example, if the size of the input image is $640 \times 640$, the sizes of the three output feature maps are $20 \times 20$, $40 \times 40$, and $80 \times 80$ respectively. Each pixel in the feature maps is fixed with three anchors of different sizes to get the intermediate results, including the probability when there is an object in each anchor $(Q, 1)$, the confidence vector of the object category $(Q, N)$, and the off-set value vector between the real object and the fixed anchor $(Q, 4)$. Therefore, there are a total of $Q = 3 \times (20 \times 20+40 \times 40+80 \times 80)$ prediction BBOXes, and each has $S = 1+4+N$ prediction values. In total, the final output of YOLO v5 is a matrix with dimension $Q \times S$. However, most of the prediction results of YOLO v5 are useless for perturbation training because some BBOXes detect either only some parts of $x$ or other objects. Therefore, we need to filter those useless BBOXes to improve the efficiency of the perturbation training.

We first extract the prediction BBOXes $O_{bbox}$ as defined below:

\begin{equation}
O_{bbox} = \bm{f}(\bm{emb}(b,x_t,bbox_{real}))
\end{equation}
\noindent
where $b$ refers to the background image, $\bm{f}(\cdot)$ is the object detector (YOLO v5). $\bm{emb}(b,x_t,bbox_{real})$ is an embedding function, in which the foreground object $x_t$ is embedded into the background image $b$ according to  $bbox_{real}$ (Algorithm 1).

Two $\bm{BF}(\cdot)$ methods, S-BBOX and M-BBOX, have been proposed to filter the prediction BBOXes of YOLO v5 and extract $k$ BBOXes for the perturbation training. S-BBOX is used for TA, while M-BBOX has two modes, one is used for HA and the other for AA and NTA.

Then, we decompose the extracted $k$ BBOXes and obtain the information for the perturbation training:

\begin{equation}
V, P = \bm{split}(\bm{BF}(O_{bbox}))
\end{equation}
\noindent
where $\bm{split}(\cdot)$ is the matrix split function. The first part is the probability $P \in \mathbb{R}^{k}$ of each target box containing the object $x$ or $x'$, the second part is the confidence vector $V \in \mathbb{R}^{k \times N}$ of the object category. Finally, $V$ and $P$ are used in the four loss functions corresponding to four attack vectors to train the perturbation.

\subsubsection{S-BBOX}

The Intersection Over Union (IOU) value between two BBOXes is calculated as follows:

\begin{equation}
\bm{iou}(bbox_A,bbox_B) =\frac{\bm{area}(bbox_A) \cap \bm{area}(bbox_B)}{\bm{area}(bbox_A) \cup \bm{area}(bbox_B)}
\end{equation}
S-BBOX first filters out most of the prediction BBOXes whose detection confidence is lower than the threshold (equal to the NMS threshold of YOLO v5). Then, S-BBOX calculates the IOU value between each prediction BBOX and $bbox_{real}$ by Equ. (5). The prediction BBOX with the highest IOU value is for the perturbation training ($k = 1$ in S-BBOX).

\subsubsection{M-BBOX}

\begin{algorithm}[t]
\DontPrintSemicolon
\caption{M-BBOX}
    \KwInput{$bbox_{pred}$: YOLO v5' s result metric; \qquad \qquad $bbox_{real}$: BBOXes obtained at algorithm 1; \qquad \qquad \qquad $bbox_{anchor}$: BBOXes of fixed anchors.}
    \KwOutput{$bbox_{training}$: BBOXes for the perturbation training}
    $bbox_{offset},bbox_{conf} = \bm{extract}(bbox_{pred})$\;
    $bbox = \bm{plus}(bbox_{offset}, bbox_{anchor})$\;
    $IOUs = \bm{iou}(bbox, bbox_{real})$\;
    \If{attack is HA} 
    {
        $index = \bm{where}(IOUs>0.5)$\;
        $middle = bbox_{pred}[index]$\;
        $middle_{conf} = bbox_{conf}[index]$\;
        $index_{conf} = \bm{top}(middle_{conf},k)$\; 
        $bbox_{training} = middle[index_{conf}]$\;
    }
    \ElseIf{attack is AA or NTA}  
    {
        $index = \bm{top}(IOUs,k)$\;
        $bbox_{training} = bbox_{pred}[index]$\;
    }
\end{algorithm}

The M-BBOX defined in Algorithm 2 has two modes, a hiding mode for HA, and a non-hiding mode for AA and NTA. $\bm{extract}(\cdot)$ extracts the $bbox_{offset}$ and $bbox_{conf}$ from the $bbox_{pred}$. $\bm{where}(\cdot)$ obtain the indexes that satisfy the confidence in the brackets. $\bm{top}(A,k)$ search the indexes of prior $k$ in the ranking of vector $A$. $\bm{plus}(\cdot)$ achieve that adding the coordinate value of two input BBOXes.

Hiding Mode: When launching HA, we first calculate the IOU values between all the prediction BBOXes and the $bbox_{real}$. Then those BBOXes with IOU greater than the threshold are filtered out (Experimental results show that $0.5$ is the best threshold). Finally, the filtered BBOXes are sorted according to their object confidence, and the prior $k$ BBOXes in the ranking of confidence are used to train perturbation.

Non-Hiding Mode: When launching AA and NTA, we first calculate the IOU values as in hiding mode. Then we directly obtain the BBOXes in the top $k$ IOU values to train perturbation.

\subsection{Four Attack Vectors}

The final goal of the four attack vectors is to deceive the target DNN models like YOLO v5 as imperceptible as possible. The optimization function is defined by:

\begin{equation}
arg\min_{x,\delta\in\mathbb{R}^{h\times w\times\varepsilon}}\mathbb{E}_{b\sim{B},t\sim{T}}\bm{loss_*}(V,P,y,y')
\end{equation}
\noindent
where $B$ refers to the background collection. We use $L_2$ distance to constrain the size of perturbation to improve its imperceptibility. The $L_2$ distance loss function is:

\begin{equation}
\mathcal{L}_{dis}={\lVert x-x'\lVert}_2^2
\end{equation}

\subsubsection{Hiding Attack}

The goal of HA is to make the object detector fail to find the object. The perturbation needs to eliminate the features of $x$ as a traffic sign by the loss function as below:

\begin{equation}
\bm{loss_{H}}=\frac{c}{nk}\sum_{i=1}^n\sum_{j=1}^k(\frac{1}{1-P_i^j})+\mathcal{L}_{dis}
\end{equation}

\subsubsection{Appearance Attack}

The main goal of AA is to make the object detector misrecognize AEs that cannot be recognized by human eyes as desired objects. This attack is to train perturbation on a blank image that can be recognized by the object detector as a specified category. Thus, the minimum Euclidean distance between AEs and the blank image is not required in AA. The loss function for AA is given as:

\begin{equation}
\bm{loss_{A}}=\frac{c}{nk}\sum_{i=1}^n\sum_{j=1}^k(\frac{1}{P_i^j \times V_i^{j,y'}})
\end{equation}

\subsubsection{Non-Target Attack}

The goal of NTA is to make the object detector misrecognize AEs that belong to a certain category as some other categories. The perturbation needs to eliminate $x$'s features of the correct category, but retain $x$'s features of a traffic sign. The loss function for NTA can be defined as:

\begin{equation}
\bm{loss_{NT}}=\frac{c}{nk}\sum_{i=1}^n\sum_{j=1}^k(\frac{1}{P_i^j}+\frac{1}{1-V_i^{j,y}})+\mathcal{L}_{dis} 
\end{equation}

\subsubsection{Target Attack}

The goal of TA is to make the object detector misrecognize AEs that belong to a certain category as the target category $y'$. The perturbation needs to not only change $x$'s features as NTA, but also add the features of $y'$. Thus, it will be tricky to design the loss function in TA. The loss function can be defined as:

\begin{equation}
\bm{loss_{T}}=\frac{c}{nk}\sum_{i=1}^n\sum_{j=1}^k(\frac{1}{P_i^j}+\frac{1}{V_i^{j,y'}}+\sum_{z=1,z\neq y}^N\frac{1}{1-V_i^{j,z}})+\mathcal{L}_{dis} 
\end{equation}
\noindent
where $N$ is the number of categories that YOLO v5 can detect. The adjustable parameter $k$ is the number of target boxes that YOLO v5 extracts from each sample. In each loss function, $c$ needs to be adjusted in each training phase according to the actual environmental conditions. $n$ is the number of samples used for training. As shown in Fig.5, the larger the value of $c$ is, the greater the perturbation will be added, and hence the more robust the generated AEs will be.

\begin{figure}[t]
  \centering
  \includegraphics[width=0.8\linewidth]{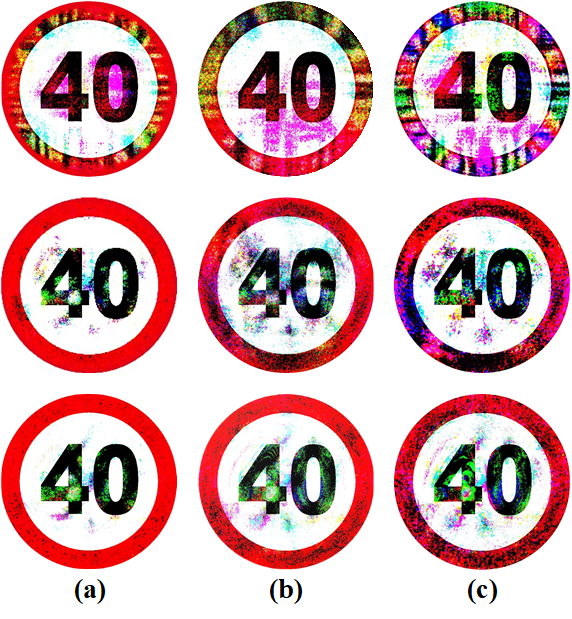}
  \caption{Impact of $c$ on imperceptibility of AEs. (a) $c$ is set to a small value. (b) $c$ is set to a medium value. (c) $c$ is set to a large value.}
\end{figure}

\section{Evaluation}

\begin{figure*}[t]
  \centering
  \includegraphics[width=0.9\linewidth]{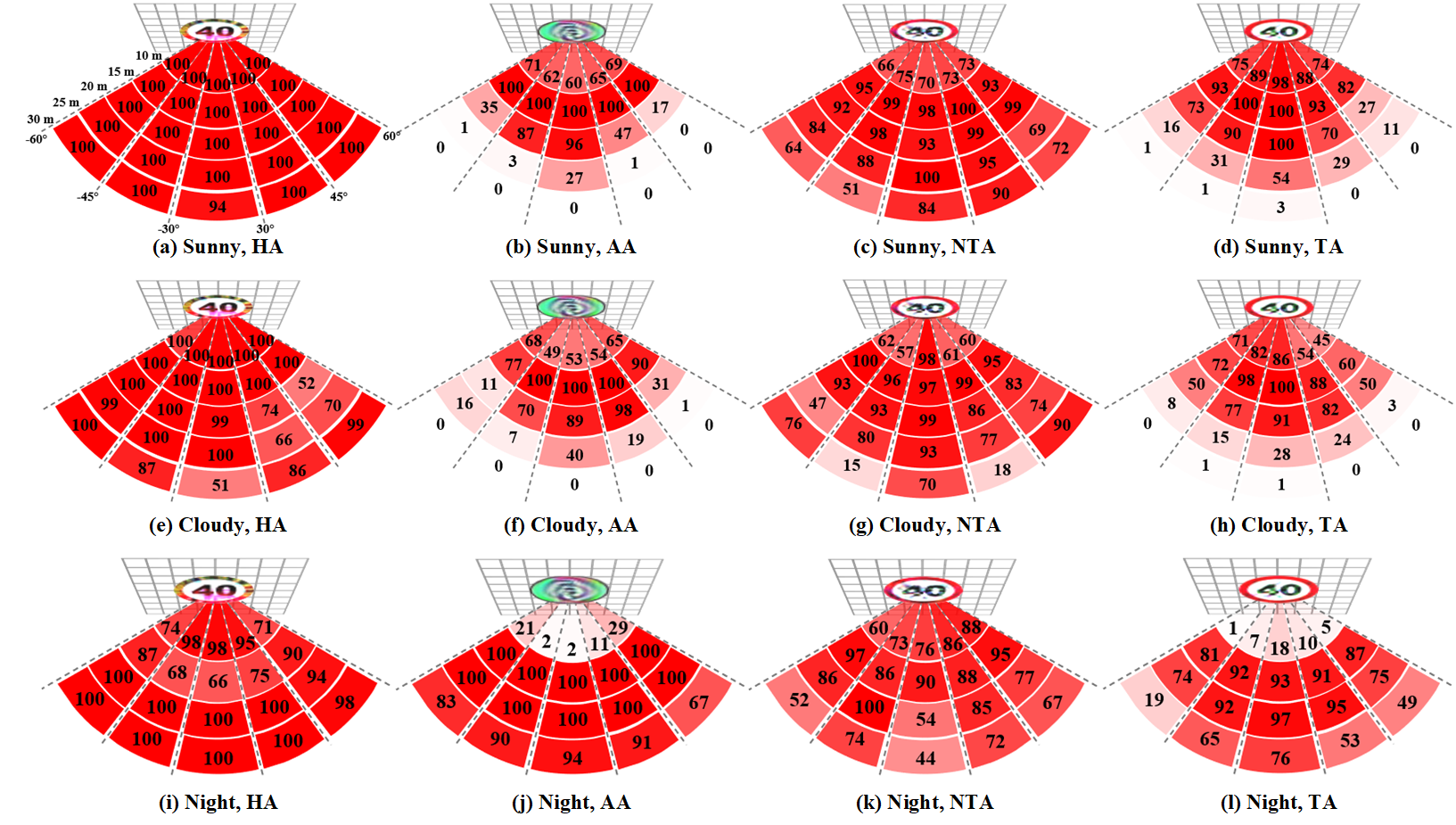}
  \caption{Attack success rate of four attack vectors under a variety of environmental conditions. (a) HA on a sunny day. (b) AA on a sunny day. (c) NTA on a sunny day. (d) TA on a sunny day. (e) HA on a cloudy day. (f) AA on a cloudy day. (g) NTA on a cloudy day. (h) TA on a cloudy day. (i) HA at night. (j) AA at night. (k) NTA at night. (l) TA at night.}
\end{figure*}

We launch the four attack vectors (\emph{i.e.}, HA, AA, NTA, and TA) against the state-of-the-art object detectors and a brand-new vehicle. Considering that the TSR system in the vehicle only recognizes the speed limit signs, we focus on physically attacking the speed limit of $40km/h$ (limit 40 for short), which is generally located on the exit ramp of the freeway thus is more life-threatening. Due to the space limit, we cannot present all the experimental results in this paper. More than 1,000 video clips and more evaluation results are uploaded on our demo website: \url{https://seczone.cn/contents/422/1024.html}.

\subsection{Experimental Setup}

AEs are generated on the server equipped with two Intel Xeon-E5-2680 V4 CPUs, four NVIDIA GTX3080 GPUs, and 64GB RECC DDR4-2400MHz memory. Then we fabricate the physical AEs in accordance with the production specifications of the traffic signs, including the material and size, \emph{etc}. We conduct multiple sets of outdoor experiments under a variety of environmental conditions. The distances range from $0m$ to $30m$, the angles range from $-60^{\circ}$ to $60^{\circ}$, and the illuminations range from sunny days, cloudy days to nights. The videos are captured by the built-in cameras of the Huawei nova7 with 64 Mega pixels and aperture of f/1.8, and the Samsung S10 with 16 Mega pixels and aperture of f/2.4. The effectiveness and robustness of the physical AEs are evaluated by analyzing each image frame in the video streaming. In order to present the threat of the physical adversarial attack more comprehensively, we not only count the attack success rate but also measure the impact of AEs on the detection confidence. 

\subsubsection{Dataset}

The YOLO v5 based TSR system is trained with the TT100k dataset \cite{R30}, which is composed of $2048 \times 2048$ images. The TT100k dataset declares that there are 221 categories containing 100,000 images of 30,000 traffic sign samples. However, we only manage to obtain 16,823 images with 26,337 traffic sign samples, 6,107 images for training, 3,073 for testing, and 7,643 other images from its website. We use the original TT100K dataset to train the object detectors first. The detectors will not converge when the high-resolution images are directly resized into the $640 \times 640$ input images for training. Thus, we divide each  $2048 \times 2048$ image into 16 $640 \times 640$ images to train the TSR system. Besides, we find that only 150 categories have data in the TT100k dataset, and among the nonempty categories, half of them have less than 10 images. Such unbalanced distribution of training data will severely impact the performance of the DNNs models \cite{R38}. After many training experiments, we screened out the traffic sign categories with top 50 data volume to form the new dataset for our experiments. The new TT100k dataset has 50 categories of 21,881 images with $640 \times 640$, 14474 images for training, 7,407 images for testing, and contains 40,550 traffic sign samples.

\begin{table}[t]
    \setlength{\abovecaptionskip}{0.1cm} 
	\newcommand{\tabincell}[2]{\begin{tabular}{@{}#1@{}}#2\end{tabular}}
	\footnotesize
	\centering
	\caption{mAP / \% of state-of-the-art object detectors.}
	\begin{tabular}{ccccc}
		\hline\hline \\[-2mm]			
		\bf{Detector} & \bf{mAP} \\ [1mm]
		\hline	 \\[-2mm]	
		\bf{Faster R-CNN (ResNet-50) \cite{R22}} & 72.6 \\ [1mm]
		\bf{SSD (VGG-16) \cite{R39}}         & 72.9  \\ [1mm]
	    \bf{RetinaNet (ResNet-50) \cite{R40}}   & 69.1 \\ [1mm]
	    \bf{CenterNet (Resnet-50) \cite{R41}}   & 72    \\ [1mm]
	    \bf{YOLO v3 (Darknet-53) \cite{R25}}     & 73.1  \\ [1mm]
	    \bf{YOLO v5 (CSPDarknet-53) \cite{R26}}     & 77.8   \\ [1mm]
		\hline\hline		
	\end{tabular}
	\vspace{0cm}
\end{table}

Table I shows the performance of the object detectors on the new dataset. Among the six detectors, Faster R-CNN \cite{R22} is a two-stage detection, others are all one-stage target detectors; CenterNet \cite{R41} is an anchor-free detector, others are all anchor-base detectors. The one-stage detectors, YOLO v3 \cite{R25}, YOLO v5 \cite{R26}, and SSD \cite{R39}, have high detection accuracy, even outperforming the two-stage detector Faster R-CNN \cite{R22}. CenterNet \cite{R41} has lower detection accuracy than most anchor-base detectors on the new dataset. In general, YOLO v5 \cite{R26} performs the best on the new dataset.

\subsubsection{Hyper-Parameters Setup}
Stochastic Gradient Descent (SGD) algorithm \cite{R42} is utilized for the optimization of the loss functions, in which the transformation batch is set to 16, and the size of the input image and the background image is set to $640 \times 640$. For the four attack vectors, we set different experimental hyperparameters as follows:

\textbf{Hiding Attack.} HA aims to hide limit 40 in the background. In HA, the $c$ is set to $1e+2$, the learning rate is set to $0.1$, and k is set to $10$.

\textbf{Appearance Attack.} AA aims to create a traffic sign which can be recognized as limit 5 by YOLO v5 from a blank image. In AA, $x$ is a randomly sampled metric with $640 \times 640$. $c$ is set to $1e+5$, the learning rate is set to $0.2$, and k is set to $1$.

\textbf{Non-target Attack.} NTA aims to make YOLO v5 recognize the limit 40 as other categories. In NTA, $c$ is set to $1e+5$, the learning rate is set to $0.1$, and k is set to $3$.

\textbf{Target Attack.} TA aims to make YOLO v5 recognize the limit 40 as the limit 60. In TA, the $c$ is set to $1e+3$, the learning rate is set to $0.1$, and k is set to $1$.

\subsubsection{Evaluation Metrics}
\begin{equation}
\begin{aligned}
N_{s} &= 
\begin{cases}
\bm{Z}(x) > th \cap \bm{Z}(x') < th, & HA \\
\bm{g}(V_{x'}) = y', & AA \\
\bm{g}(V_{x}) = y \cap \bm{g}(V_{x'}) \neq y,& NTA \\
\bm{g}(V_{x}) = y \cap \bm{g}(V_{x'}) = y',& TA \\
\end{cases}
\\
N_{a} &= 
\begin{cases}
1,& HA \& AA \\
\bm{g}(V_{x}) = y,& TA \& NTA \\
\end{cases}
\end{aligned}
\end{equation}

\begin{figure}[t]
  \centering
  \includegraphics[width=\linewidth]{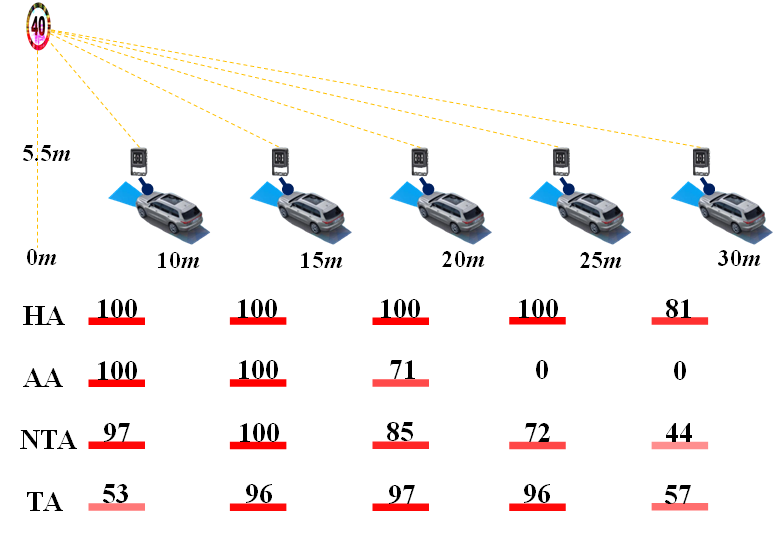}
  \caption{Impact of height on attack effectiveness.}
\end{figure}

The attack success rate is defined as $R_s = \sum N_s/\sum N_a \times 100\%$. As shown in Equ. (12), $\bm{g}(\cdot)$ denotes the $\bm{argmax}(\cdot)$ function that outputs the index of the max value in the vector. In HA and AA, $N_a$ is every image frame captured by the camera, while in NTA and TA, $N_a$ is the image frame in which the object is correctly detected. $N_s$ depends on the attack goals of the four attack vectors. In HA, a successful attack means that the probability of detecting $x$ ($\bm{Z}(x) = \bm{\max}(V_{x}) \times P_{x}$) is greater than the threshold $th=0.25$, and the probability of detecting $x'$ ($\bm{Z}(x') = \bm{\max}(V_{x'}) \times P_{x'}$) is less than the threshold. In AA, a successful attack means $x'$ is detected as the target category $y'$. In NTA, a successful attack means $x$ is detected as $y$, but $x'$ is not detected as the original category $y$. In TA, a successful attack means $x$ is detected as $y$, and $x'$ is detected as the target category $y'$.

\begin{figure*}[p]
  \centering
  \includegraphics[width=0.9\linewidth]{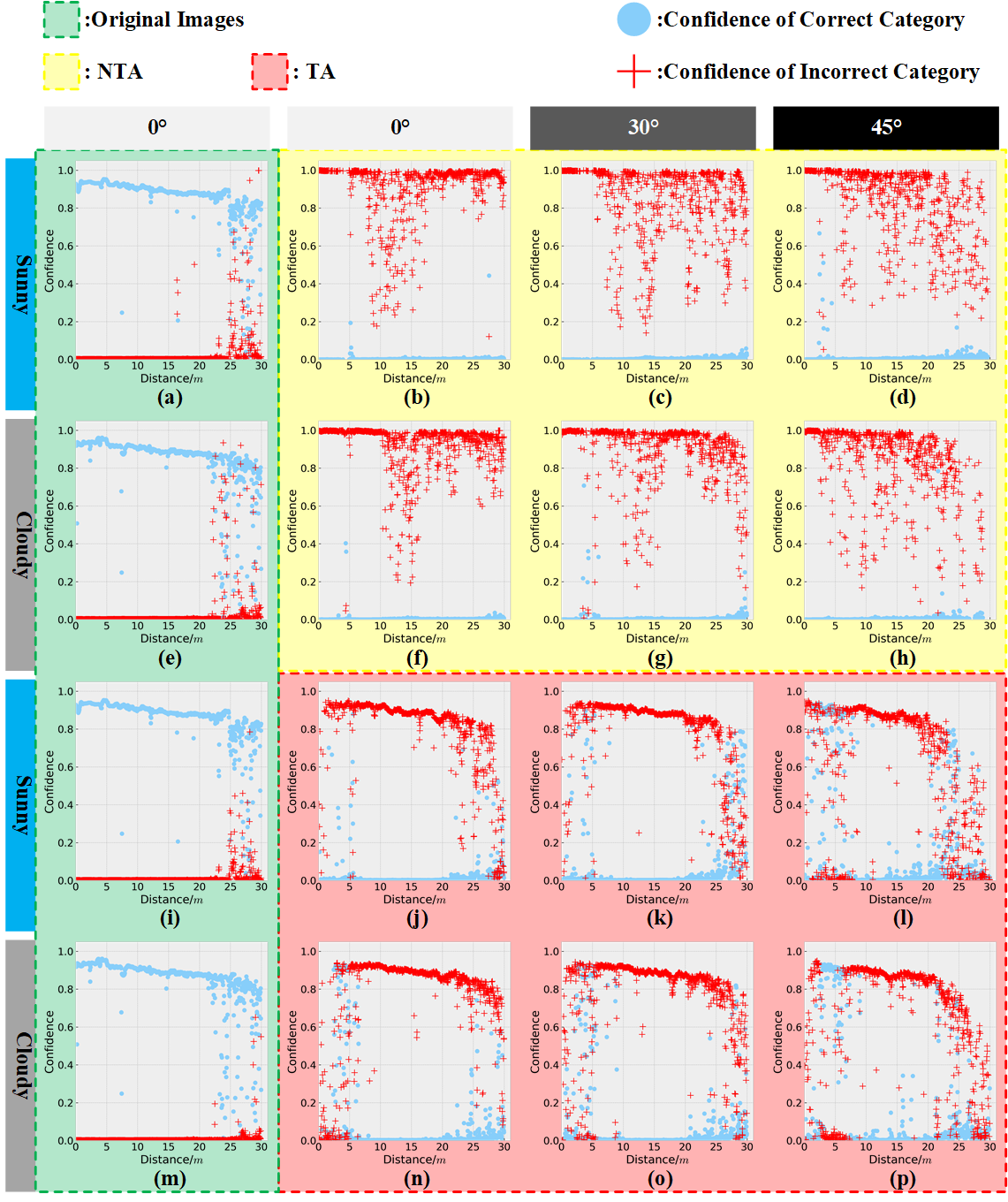}
  \caption{Confidences of NTA and TA under a variety of environmental conditions. (a) NTA with original image on a sunny day. (b) NTA at $0^{\circ}$ on a sunny day. (c) NTA at $30^{\circ}$ on a sunny day. (d) NTA at $45^{\circ}$ on a sunny day. (e) NTA with original image on a cloudy day. (f) NTA at $0^{\circ}$ on a cloudy day. (g) NTA at $30^{\circ}$ on a cloudy day. (h) NTA at $45^{\circ}$ on a cloudy day. (i) TA with original image on a sunny day. (j) TA at $0^{\circ}$ on a sunny day. (k) TA at $30^{\circ}$ on a sunny day. (l) TA at $45^{\circ}$ on a sunny day. (m) TA with original image on a cloudy day. (n) TA at $0^{\circ}$ on a cloudy day. (o) TA at $30^{\circ}$ on a cloudy day. (p) TA at $45^{\circ}$ on a cloudy day.}
\end{figure*}

\subsection{Evaluation of Four Attack Vectors}

\subsubsection{Attack Success Rate}
As shown in Fig.6, the four attack vectors are evaluated in terms of the attack success rate $R_s$ under a variety of environmental conditions. In this set of experiments, we fixed the height of the traffic signs to about $1.5m$. The distances are divided into five regions, \emph{i.e.}, $[0m, 10m]$, $[10m, 15m]$,  $[15m, 20m]$, $[20m, 25m]$, $[25m, 30m]$. The video clips are recorded from far to close with a constant speed in each region at the angle of $[-60^{\circ}, -45^{\circ}]$, $[-45^{\circ}, -30^{\circ}]$, $[-30^{\circ}, 30^{\circ}]$, $[30^{\circ}, 45^{\circ}]$, and $[45^{\circ}, 60^{\circ}]$, respectively. For each $R_s$ in Fig.6, more than 200 image frames are captured and processed (Fig.13 and Fig.14 in appendix). Since the visibility is low at night, we only conduct experiments within $25m$. The depth of red represents the value of $R_s$, the darker the color indicates the higher the attack success rate.

For HA, we achieve the best attack effectiveness on a sunny day. On a cloudy day, in the region of $[15m, 20m]$ at $[45^{\circ}, 60^{\circ}]$ and in the region of $[25 m, 30m]$ at $[-30^{\circ}, 30^{\circ}]$, $R_s$ degrades to 50\%. At night, in the region of $[10m, 15m]$ at $[-45^{\circ}, -30^{\circ}]$ and $[-30^{\circ}, 30^{\circ}]$, $R_s$ degrades below 70\%. However, HA has a high $R_s$ at $10m$ away in the dark environment. The experimental results indicated that reflection at night has a greater impact on HA.

For AA, we achieve the best attack effectiveness at night, especially in the region of $[10m, 20m]$. However, in the dark environment within $10m$, $R_s$ is very low. On sunny and cloudy days, $R_s$ is high in the region of $[10m, 15m]$, but AA almost failed at $25m$ away. Thus, $R_s$ is relatively higher in the darker light, while both larger angles and distances will significantly reduce $R_s$. The experimental results indicated the huge impact of reflection at night on AA too.

For NTA, since the strong reflection caused by direct lighting covers the perturbation features, $R_s$ degrades below 60\% in the region of $[15m, 20m]$ at $[-30^{\circ}, 30^{\circ}]$ at night. On a cloudy day, $R_s$ degrades below 20\% in the region of $[25m, 30m]$ at $[-45^{\circ}, -30^{\circ}]$ and $[30^{\circ}, 45^{\circ}]$. The results indicate that both darker light and larger distances reduce $R_s$, while the impact of different angles on $R_s$ does not show a strong regularity. 

For TA, we achieve the best attack effectiveness in the region of $[10m, 20m]$ at $[-45^{\circ}, 45^{\circ}]$. At night, TA almost fails within $10m$ in the dark environment. On a sunny day and a cloudy day, $R_s$ degrades severely as the distance increases at $20m$ away. The results indicate that large angles reduce $R_s$ in TA when the distance is either very close or very far; while the impact of different illuminations on $R_s$ does not show a strong regularity. 

Then we increased the height of the traffic sign to about $5.5m$ and conducted the four attack vectors on a sunny day at $0^{\circ}$ as shown in Fig.11 in Appendix. 

The experimental results in Fig.7 indicate that the height of AEs has little impact on HA within $25m$, but $R_s$ drops by 13\% in the region of $[25m, 30m]$. For AA, $R_s$ rises from 60\% to 100\% in the region of $[0m, 10m]$ when the height of AEs rises from $1.5m$ to $5.5m$, but $R_s$ drops by 25\% in the region of $[15m, 20m]$. Thus, $R_s$ in AA increases with height in the closer region, while at $20m$ away, $R_s$ in AA decreases as the height of AEs increases. Similarly, NTA has higher $R_s$ in the region of $[0m, 20m]$ and lower $R_s$ at $20m$ away when the height of AEs rises from $1.5m$ to $5.5m$. For TA, $R_s$ drops by 45\% in the region of $[0m, 10m]$, but rises from 54\% to 96\% in the region of $[20m, 25m]$ and rises from 3\% to 57\% in the region of $[25m, 30m]$ when the height of AEs rises from $1.5m$ to $5.5m$. 

We also test the four attack vectors in a vehicle with a speed of $[20km/h, 30km/h]$ as shown in Fig.12 in Appendix. The test is conducted in the region of $[0m, 30m]$ at $[-30^{\circ}, 30^{\circ}]$ on a sunny day. The attack effectiveness is listed in Table II. The average $R_s$s in HA, NTA, and TA are all over 90\%. The results of the real-road driving test for AA are consistent with the results in Fig.6. The experimental results in Table II show that our generated AEs are robust in the real-road driving test.

\begin{table}[t]
    \setlength{\abovecaptionskip}{0.1cm} 
	\newcommand{\tabincell}[2]{\begin{tabular}{@{}#1@{}}#2\end{tabular}}
	\footnotesize
	\centering
	\caption{Real-road driving test}
	\begin{tabular}{ccc}
		\hline\hline \\[-2mm]			
		\bf{Attack Vector} & \bf{Number of image frames} & $\bm{R_s/ \%}$ \\ [1mm]
		\hline	 \\[-2mm]	
		\bf{HA}   & 824 & 96.48 \\ [1mm]
		\bf{AA}   & 630 & 60.48 \\ [1mm]
	    \bf{NTA}  & 525 & 90.48 \\ [1mm]
	    \bf{TA}   & 645 & 92.87  \\ [1mm]
		\hline\hline		
	\end{tabular}
	\vspace{0cm}
\end{table}

\begin{table}[t]
    \setlength{\abovecaptionskip}{0.1cm} 
	\newcommand{\tabincell}[2]{\begin{tabular}{@{}#1@{}}#2\end{tabular}}
	\footnotesize
	\centering
	\caption{$R_s$ / \% and mAP / \% on state-of-the-art object detectors}
	\begin{tabular}{ccccc}
		\hline\hline \\[-2mm]			
		 & \tabincell{c}{\bf{Faster}\\\bf{R-CNN\cite{R22}}} & \bf{SSD\cite{R39}} & \bf{RetinaNet\cite{R40}} & \bf{CenterNet\cite{R41}}\\ [1mm] 
		\hline	 \\[-2mm]	
		\bf{HA}   & 95.02 & 95.40 & 71.02 & 97.64   \\ [1mm]
		\bf{AA}   & 54.73 & 28.23 & 11.94 & 19.65   \\ [1mm]
	    \bf{NTA}  & 99.78 & 100   & 52.18 & 47.8 \\ [1mm]
	    \bf{TA}   & 58.62 & 46.38 & 2.03  & 0  \\ [1mm]
	    \hline   \\[-2mm]
	    \bf{mAP-s} & 61.5 & 64.6 & 48.5 & 55.3  \\ [1mm]
	    \bf{mAP-m} & 77.7 & 79.3 & 77.2 & 77.9  \\ [1mm]
	    \bf{mAP-l} & 82.5 & 83.7 & 84 & 85.7  \\ [1mm]
		\hline\hline		
	\end{tabular}
	\vspace{-0.3cm}
\end{table}

\subsubsection{Confidence}

In order to further study the effectiveness and robustness of NTA and TA, we measure the detection confidence of each image frame. We fix the height of the traffic signs to about $1.5m$ and choose the video clips recorded on a sunny day and a cloudy day in the region of  $[0 m, 30 m]$ at $0^{\circ}$, $30^{\circ}$, and $45^{\circ}$, respectively. As shown in Fig.8, the blue circle presents confidence of the correct category $y$. It needs to be emphasized that in NTA, the red cross represents the highest confidence of all the incorrect categories, while in TA, it represents the confidence of the target category $y'$.   

Fig.8(a,e,i,m) in the leftmost column show confidence in each original input image. Limited by the resolution of the camera, the detection performance of the original images drops significantly at $25 m$ away. In several image frames, the confidence of the correct category $y$ drops severely, while the confidence of the incorrect categories increases. 

By comparing the results of NTA on a sunny day in Fig.8(b,c,d) and the results of NTA on a cloudy day in Fig.8(f,g,h), we find that the illumination has little impact on the attack effectiveness of NTA. By comparing the results of TA on a sunny day in Fig.8(j,k,l) and the results of TA on a cloudy day in Fig.8(n,o,p), it indicates that the attack effectiveness of TA is better in a sunny day within $10m$. By comparing the experimental results at $0^{\circ}$ in Fig.8(b,f,j,n), $30^{\circ}$ in Fig.8(c,g,k,o), and $45^{\circ}$ in Fig.8(d,h,l,p), it indicates three phenomena. First, the large angle has a greater impact on TA than NTA. Second, at very close and far distances, the large angle significantly degrades the attack effectiveness of TA. Third, attack effectiveness is better in a sunny day than in a cloudy day.

\begin{figure*}[t]
  \centering
  \includegraphics[width=0.9\linewidth]{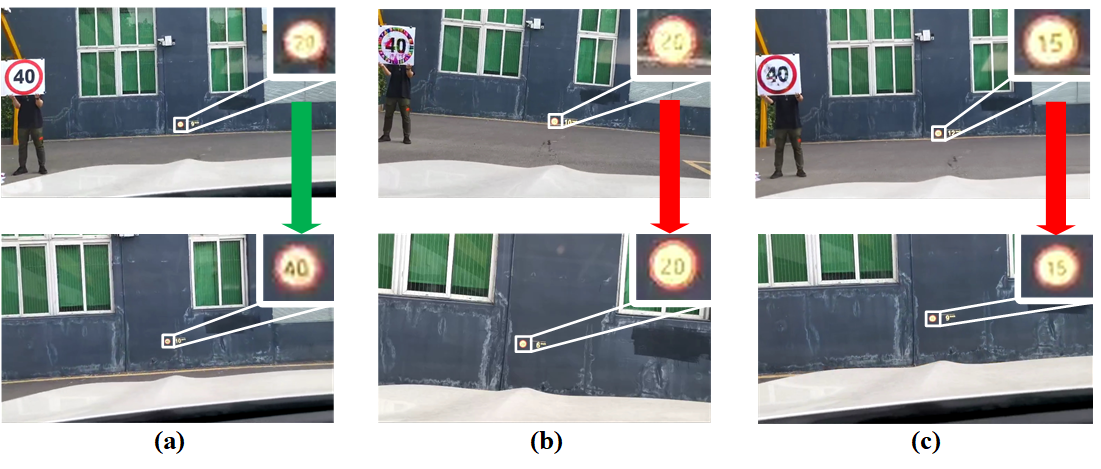}
  \caption{HA and NTA against a brand-new vehicle. (a) The original speed limit 40. (b) HA. (c) NTA.}
\end{figure*}

\subsection{Attack Transferability}
\subsubsection{State-of-the-art Object Detectors}
We use AEs generated with YOLO v5 to attack the representative object detectors, including Faster R-CNN (two-state) \cite{R22}, SSD \cite{R39} and RetinaNet \cite{R40} (one-stage and anchor-base), and CenterNet (one-stage and anchor-free) \cite{R41}. Those object detectors are trained on the same TT100k \cite{R30} dataset for fairness. The set of experiments is conducted in the region of $[0m, 30m]$ at $0^{\circ}$. According to the experimental results in Table III, HA and NTA achieve higher $R_s$, especially on SSD and Faster R-CNN, while AA and TA on RetinaNet and CenterNet almost fail. We measure the performance of the four detectors when detecting large, medium and small objects, and find that RetinaNet and Centernet have low accuracy when detecting small objects. Therefore, low accuracy at long distances determines that they have low transferability in the region of [$20m$, $30m$], but they still have high accuracy at close distances, and thus have good transferability in the region of [$0m$, $15m$].

\subsubsection{Brand-new Vehicle}
The major reason why we generated adversarial limit 40 is that the TSR system of the brand-new vehicle can only recognize the speed limit signs. As shown in Fig.9(a), only after the vehicle passes the speed limit sign, the recognition result can be displayed on the screen. In HA and NTA, Fig.9(b,c) show that after the vehicle passes the adversarial limit 40, it does not display the correct limit 40, which means HA and NTA successfully fool the eyes of the vehicle.

\begin{table*}[t]
    \setlength{\abovecaptionskip}{0.1cm} 
	\newcommand{\tabincell}[2]{\begin{tabular}{@{}#1@{}}#2\end{tabular}}
	\footnotesize
	\centering
	\caption{Comparison of different adversarial attack methods in physical domain}
	\begin{tabular}{ccccc}
		\hline\hline \\[-2mm]			
		 & \bf{Our Method} & \bf{Zhao's Method \cite{R18}} & \bf{ShapeShifter \cite{R31}} & \bf{NaturalAE \cite{R37}}\\ [1mm] 
		\hline	 \\[-2mm]
		\bf{Target Model}   & YOLO v5 & YOLO v3 & Faster R-CNN & YOLO v2  \\ [1mm]
		\bf{Attack Vector}   & HA, AA, NTA, TA & HA, AA & TA & TA   \\ [1mm]
		\bf{Distance}   & [0 m, 30 m] & [0 m, 25 m] & [0 m, 12 m] & [0 m, 5 m]   \\ [1mm]
	    \bf{Angle}  & [-60,60]   & [-60,60] & [0,60] & [-60,60] \\ [1mm]
	    \bf{Illumination}   & Sunny, Cloudy, Night & Indoor, Outdoor  & Indoor, Outdoor & Indoor, Outdoor  \\ [1mm]
	    \bf{Height}   & 1.5 m, 5.5 m & 1.5m  & 1.5m & 1.5m  \\ [1mm]
	    \bf{Transferability}   & \tabincell{c}{Faster R-CNN, SSD, RetinaNet, CenterNet\\Brand-new vehicle (HA and NTA)} & \tabincell{c}{Faster R-CNN, SSD\\RFCN, Mask R-CNN} & None & Faster R-CNN, SSD  \\ [1mm]
		\hline\hline		
	\end{tabular}
	\vspace{-0.3cm}
\end{table*}

\subsection{Comparison and Discussion}

There are few works focusing on the physical AEs against the object detectors. As listed in Table IV, We compare our proposed method with Zhao's method \cite{R18}, ShapeShifter \cite{R31}, and NaturalAE \cite{R37} from different dimensions. In general, our method has the following advantages. First of all, to the best of our knowledge, it is the first attempt to physically attack a production-grade object detector in a real-world vehicle and achieve significant effectiveness. Second, we take into account the impact of light reflection in the dark environment on robustness and conduct a complete set of experiments at night. Surprisingly, all the four attack vectors achieve almost 100\% $R_s$ against the YOLO v5 based object detector at a certain distance and angle. Third, NTA and TA are implemented in the physical domain and exhibit satisfactory transferability against the state-of-the-art object detectors, which lays a foundation for studying the adversarial attacks against black-box models in both the digital and the physical domains.

Combining all the experimental results, we can draw a conclusion that the physical AEs for HA and NTA have better robustness and transferability against multiple object detectors, while AA and TA have high $R_s$ only on specific attack targets. The lower $R_s$ and poorer transferability are due to two-fold reasons. First, HA only needs to eliminate the features of all the categories in the perturbation training process, and NTA only needs to eliminate the features of the correct category $y$. But for AA and TA, the features of the target category $y'$ should be imitated, which is much more difficult than the feature elimination. In addition, TA also needs to eliminate the features of $y$ before the feature imitation. Therefore, the digital-physical-digital conversion has a greater impact on the AEs generated for AA and TA, which results in lower transferability. Second, the similarity of the two target models also determines the transferability of the physical AEs. The architectures of RetinaNet and CenterNet are quite different from YOLO v5. As for the TSR system of the brand-new vehicle, even the architecture of the DNNs model is unknown. The black-box setting is always the biggest obstacle for the adversarial attack in both digital and physical domains. In future work, we will launch the physical adversarial attack against the black-box DNNs model with the help of the side-channel attack \cite{R43}. 

\section{Related Work}

We survey the landmark works related to the adversarial attacks in Fig.10, and discuss the AEs defense mechanisms to protect the DNNs models from the adversarial attacks.

\subsection{Adversarial Attacks}

\begin{figure}[t]
  \centering
  \includegraphics[width=0.85\linewidth]{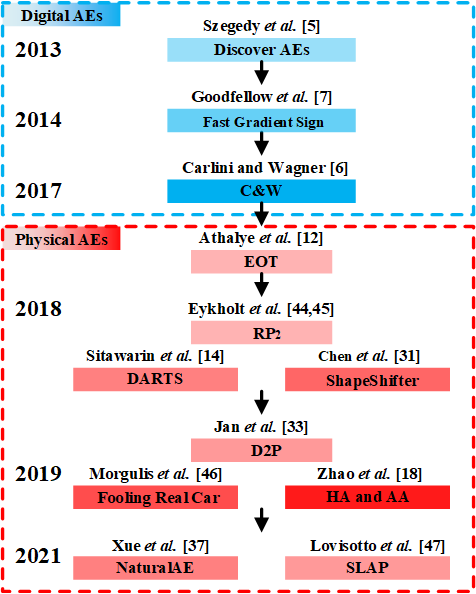}
  \caption{Adversarial Attacks.}
\end{figure}

\subsubsection{Digital AEs}

In 2013, Szegedy \emph{et al.} \cite{R5} discovered the existence of AEs in the digital domain and generated AEs using the box-constrained L-BFGS. Since then, there have been a lot of initial works \cite{R6,R7,R8,R9,R10} focusing on generating digital AEs. Goodfellow \emph{et al.} \cite{R7} proposed the fast gradient sign method to generate AEs on the MNIST dataset. Carlini and Wagner \cite{R6} introduced three C\&W attacks for the $L_0$, $L_2$, and $L_{\infty}$ distance metrics. The $L_0$ attack was the first published attack that caused the targeted misclassification on the ImageNet dataset. Although most of the digital AEs lose their adversarial nature in the physical environment \cite{R35,R36}, these three classic methods are still used for generating physical AEs.

\subsubsection{Physical AEs}

Athalye \emph{et al.} \cite{R12} proposed the Expectation Over Transformations (EOT) method that laid the cornerstone of the physical AEs generation algorithms. Evtimov \emph{et al.} \cite{R44,R45} used EOT method to generate robust physical AEs to attack YOLO v2 based TSR systems. Sitawarin \emph{et al.} \cite{R14} presented two novel attack vectors against the traffic sign classifier. Based on \cite{R14}, Morgulis \emph{et al.} \cite{R46} claimed to attack the real production-grade image classifiers for the first time. Chen \emph{et al.} \cite{R31} physically attacked the fast R-CNN based TSR systems using large perturbations. Lovisotto \emph{et al.} \cite{R47} used the light of a projector to generate short-lived adversarial perturbations in the indoor experiments. However, the physical AEs generated by these methods cannot remain robust under a variety of environmental conditions.

Until very recently, there were several attempts to improve the robustness of the physical AEs. Jan \emph{et al.} \cite{R33} used an image-to-image translation network to simulate the digital-physical conversion for generating the physical AEs. This improvement performed well on several image classifiers with static input images, but it was not verified on the object detectors with video streaming. Xue \emph{et al.} \cite{R37} proposed the natural and robust physical AEs against the object detectors. Since the generated AEs are effective within 5 meters, they pose almost no threat to a moving vehicle. Zhao \emph{et al.} \cite{R18} presented the hiding attack and appearance attack on the YOLO v3 based object detector and achieved better robustness. Limited by the capability of the AEs generation algorithm, the imperceptibility of AEs is not properly addressed which makes the two attack vectors lack real-world threat to the production-grade TSR systems.

\subsection{Defense Mechanisms}

\begin{figure*}[t]
  \centering
  \includegraphics[width=\linewidth]{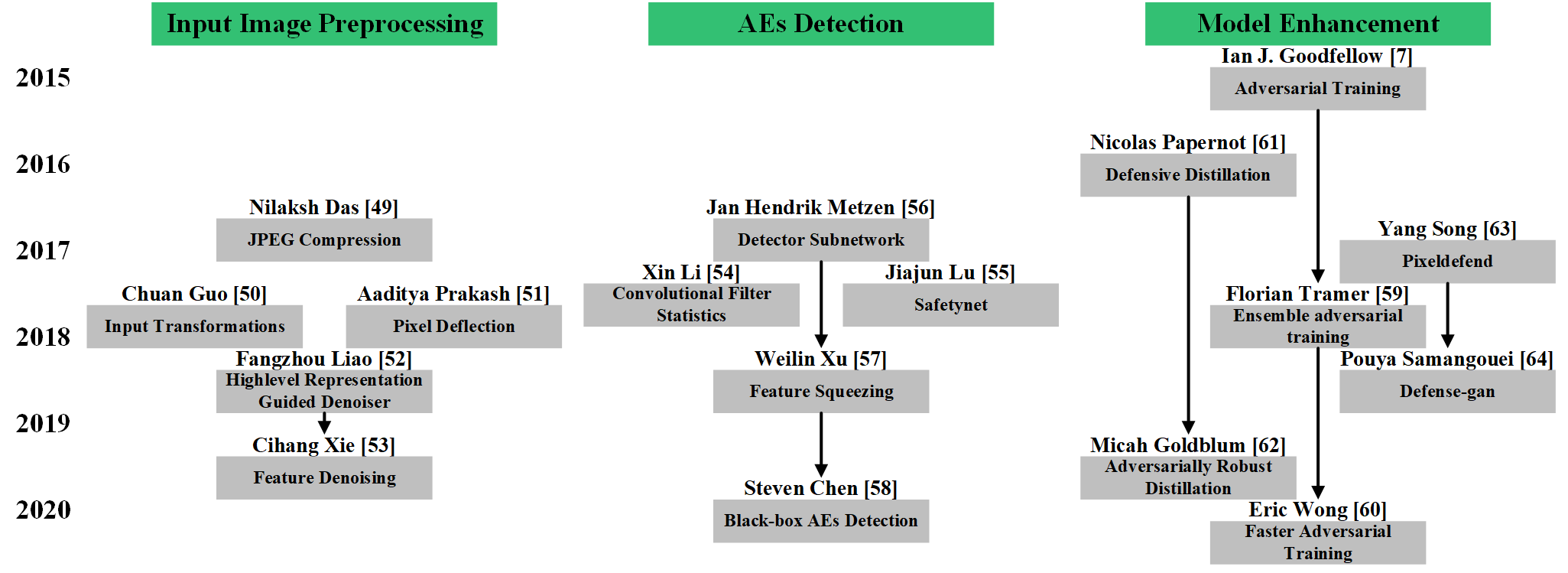}
  \caption{Defense mechanisms against AEs. (a) Input image preprocessing. (b) AEs detection. (c) Model enhancement. }
\end{figure*}

Research has proposed several defense mechanisms to protect the DNNs models against AEs, which can be divided into three categories,\emph{i.e.}, input image preprocessing, AEs detection, and model enhancement. 

\subsubsection{Input Image Preprocessing}

Imperceptible requirements make AEs not robust enough to the external noises or data distortions \cite{R48}. Inspired by this opportunity, preprocessing the input image to eliminate the adversarial features could be a type of potential defense mechanism. In 2017, Das \emph{et al.} \cite{R49} explored and demonstrated that the system's JPEG compression could be used as an effective preprocessing step in the classification pipeline to significantly reduce the impact of AEs. JPEG compression has ability to remove the high-frequency signal components inside the image, which helps eliminate malicious disturbances. Guo \emph{et al.} \cite{R50} selected a small group of pixels and reconstructs the simplest image consistent with the selected pixels so that malicious disturbance in the image was removed. Similar to \cite{R50}, Prakash \emph{et al.} \cite{R51} proposed a pixel deflection method that used semantic maps and randomization to select a small number of pixels, and then replace them with randomly selected neighboring pixels. This process would generate noise and needed a wavelet noise reduction filter. In order to solve this problem, Liao \emph{et al.} \cite{R52} used U-net structure to modify the denoising autoencoder and proposed denoising U-net. However, the residual noise impact might still increase as the number of network layers increases. In 2019, Xie \emph{et al.} \cite{R53} developed a new network architecture to improve robustness with a feature denoiser that combined confrontation training and graphics processing.

\subsubsection{AEs Detection}

It is also possible to directly detect the adversarial features and prevent AEs from entering DNNs models. Li \emph{et al.} \cite{R54} used a cascade classifier to detect AEs effectively. Lu \emph{et al.} \cite{R55} proposed SafetyNet to detect AEs based on that benign images and AEs have different ReLU activation patterns in the model. Metzen \emph{et al.} \cite{R56} chose a small detector sub-network to enhance the robustness of the DNNs model. The sub-network would be trained on a binary classification task to distinguish AEs from benign images. On this basis, Xu \emph{et al.} \cite{R57} proposed a feature compression method to detect AEs by comparing the prediction consistency of the original image and the compressed image. If the classification result of the original input image and the compressed input image were different, the input might be an AE. In 2020, Chen \emph{et al.} \cite{R58} presented a defense mechanism to detect the process of AEs generation. By keeping the historical record of past queries, it could determine when a series of queries appeared to generate AEs. 

\subsubsection{Model Enhancement}

Enhancing the DNNs models in the training phase is also a potential defense mechanism against adversarial attacks. In the next year after the AEs concept appeared, Goodfellow \emph{et al.} \cite{R7} proposed to add AEs to the training set to make the trained DNNs model robust against specific types of AEs. Tramer \emph{et al.} \cite{R59} believed that \cite{R7} was vulnerable to single-step attacks. They proposed an ensemble adversarial training method based on gradient masks, using AEs generated by other pre-trained classifiers to expand the training set. This ensemble adversarial training algorithm was more effective because it separated the training of the model and the process of generating AEs. In 2020, Wong \emph{et al.} \cite{R60} paid more attention to the cost of adversarial training, and proposed a fast adversarial training method. Defensive distillation was proposed by Papernot \emph{et al.} \cite{R61} in 2016. The basic idea was to transfer knowledge of complex networks to simple networks by modifying the network structure and optimization items to prevent the model from fitting too closely to normal samples. However, the network needed to be retrained and was usually only effective against AEs considered in the training process. In 2020, Goldblum \emph{et al.} \cite{R62} studied how the adversarial robustness is transferred from the teacher DNNs model to the student DNNs model in the knowledge distillation process, and introduced adversarial robust distillation to refine the robustness to the student DNNs. Another robustness enhancement method was to use a generative model to project potential AEs onto a benign dataset, and then classify them. Among them, the PixelDefend method proposed by Song \emph{et al.} \cite{R63} used the PixelCNN generative model, and the Defence-GAN method proposed by Samangouei \emph{et al.} \cite{R64} used a generative adversarial network structure.

\section{Conclusion}

In this paper, we present a systematic pipeline to generate the physical AEs against the object detectors. In order to improve the robustness of AEs in the physical domain, we extend the distribution of image transformations, design the S-BBOX filter and the M-BBOX filter, and modify the four loss functions for the four attack vectors respectively. We conduct extensive experiments under a variety of environmental conditions, \emph{i.e.}, the distance varies from $0m$ to $30m$, the angle varies from $-60^{\circ}$ to $60^{\circ}$, the illumination varies from sunny day to cloudy day to night. The experimental results show that HA, NTA, and TA achieve a success rate of more than 90\% in real-world driving tests against the YOLO v5 based TSR system. The generated AEs exhibit high transferability against the other state-of-the-art object detectors. Furthermore, HA and NTA successfully fooled the TSR system of a brand-new vehicle, which is a life-threatening case for autonomous vehicles.

\section*{Acknowledgment}
First, the authors would like to thank Mr. Haoxiang Yang and Mr. Tiancheng Ji, who helped to conduct the on-road experiments in the hot summer. 

Zhaojun Lu was partially supported by the National Natural Science Foundation of China (Grant No. 61874047), a grant of key technologies R\&D general program of Shenzhen (No.JSGG20201102170601003), and the National Key Research and Development Program (2019YFB1310001).

\small
\bibliographystyle{IEEEtran}
\bibliography{V_CR2}

\onecolumn
\appendix
\section{Appendix}

\begin{table}[h]
    \setlength{\abovecaptionskip}{0.1cm} 
	\newcommand{\tabincell}[2]{\begin{tabular}{@{}#1@{}}#2\end{tabular}}
	\footnotesize
	\centering
	\caption{Definitions of the notations in image frames}
	\begin{tabular}{ccccccc}
		\hline\hline \\[-2mm]			
		\bf{Notations in Images} & pl5 & pl40 & pl60 & pl80 & pm50 & po\\ [1mm] 
		\hline	 \\[-2mm]	
		\bf{Definitions} & $5km/h$ speed limit & $40km/h$ & $60km/h$t & $80km/h$ & $55ton$ weight limit & other prohibition signs    \\ [1mm]
		\hline\hline		
	\end{tabular}
	\vspace{-0.3cm}
\end{table}

\begin{figure*}[h]
  \centering
  \includegraphics[width=\linewidth]{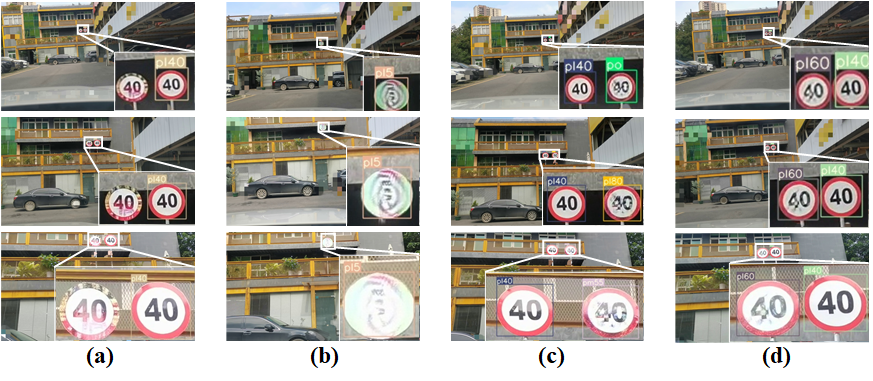}
  \caption{Results of four attack vectors at a height of $5.5m$. (a) HA. (b) AA. (c) NTA. (d) TA.}
\end{figure*}

\begin{figure*}[h]
  \centering
  \includegraphics[width=\linewidth]{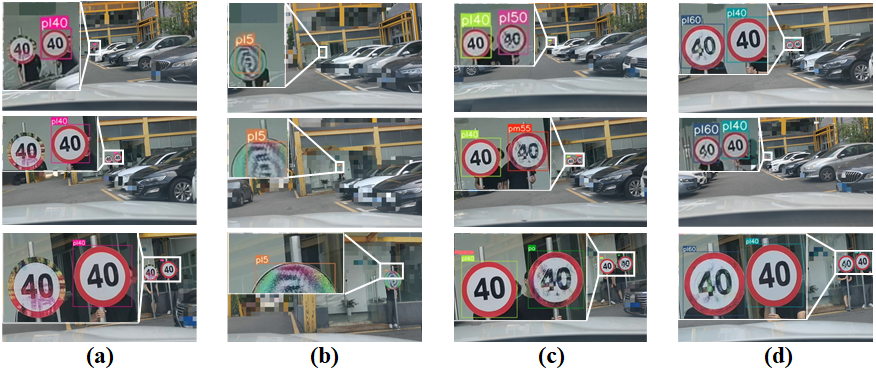}
  \caption{Results of four attack vectors in real-road driving tests. (a) HA. (b) AA. (c) NTA. (d) TA.}
\end{figure*}

\begin{figure*}[t]
  \centering
  \includegraphics[width=0.9\linewidth]{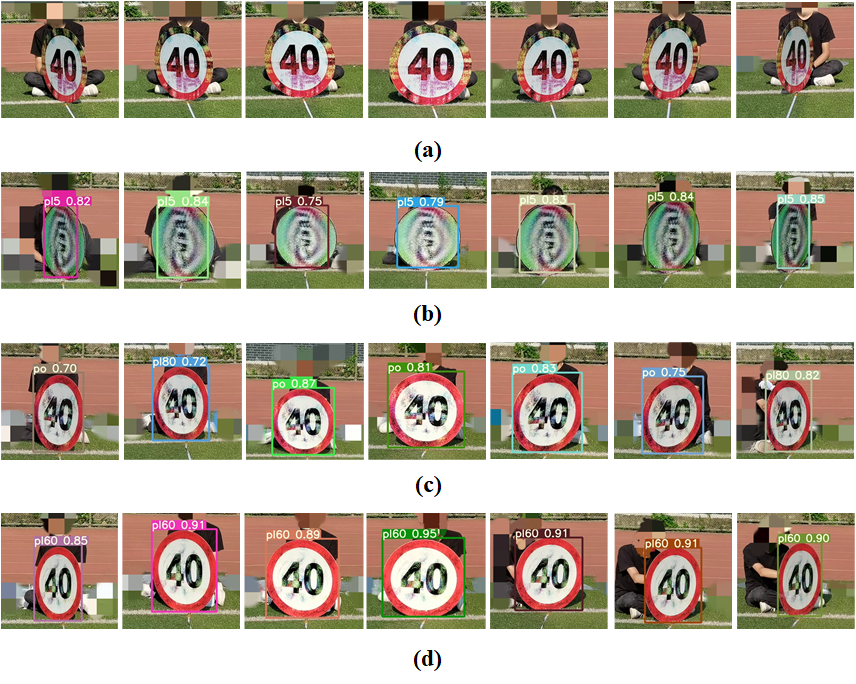}
  \caption{Results of four attack vectors at varying angles ($-60^{\circ}$,$-45^{\circ}$,$-30^{\circ}$,$0^{\circ}$,$30^{\circ}$,$45^{\circ}$,$60^{\circ}$). (a) HA. (b) AA. (c) NTA. (d) TA.}
\end{figure*}

\begin{figure*}[b]
  \centering
  \includegraphics[width=0.8\linewidth]{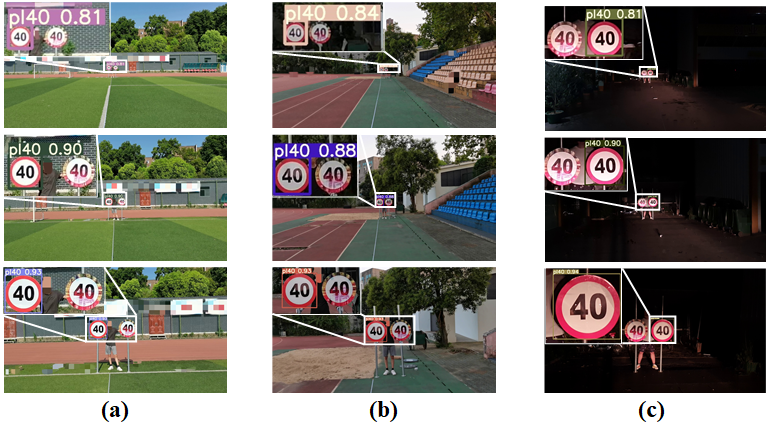}
  \caption{Results of HA under different illuminations. (a) Sunny day. (b) Cloud day. (c) Night.}
\end{figure*}

\end{document}